\documentclass[10pt,journal,compsoc]{IEEEtran}

\usepackage{amsmath}
\usepackage{amssymb}
\usepackage{latexsym}
\usepackage[linesnumbered,ruled]{algorithm2e}

\SetCommentSty{mycommfont}

\usepackage{framed}
\usepackage{tabularx}
\usepackage{multirow}
\usepackage{booktabs}
\usepackage{adjustbox}
\usepackage{graphicx}
\usepackage{xcolor}
\definecolor{newcolor}{rgb}{.8,.349,.1}

\ifCLASSOPTIONcompsoc
  \usepackage[nocompress]{cite}
\else
  \usepackage{cite}
\fi

\usepackage{hyperref}
\usepackage[capitalize]{cleveref}
\crefname{section}{Sec.}{Secs.}
\Crefname{section}{Section}{Sections}
\Crefname{table}{Table}{Tables}
\crefname{table}{Tab.}{Tabs.}
\crefname{equation}{Eq.}{Eqs.}
\crefname{algocf}{Alg.}{Algs.}

\usepackage{url}

\begin{document}
\title{CP-CNN: Core-Periphery Principle Guided Convolutional Neural Network}
%


\author{Lin Zhao, Haixing Dai, Zihao Wu, Dajiang Zhu, and Tianming Liu, ~\IEEEmembership{Senior Member,~IEEE}% <-this % stops a space
\IEEEcompsocitemizethanks{
\IEEEcompsocthanksitem L. Zhao, H. Dai, Z. Wu and T. Liu are with School of Computing, University of Georgia, Athens, GA, 30602.
E-mail: tianming.liu@gmail.com
\IEEEcompsocthanksitem D. Zhu is with Department of Computer Science and Engineering, University of Texas at Arlington, Arlington, TX, 76013.}% <-this % stops an unwanted space
\thanks{Manuscript received April 19, 2005; revised August 26, 2015.}}

\markboth{Journal of \LaTeX\ Class Files,~Vol.~14, No.~8, August~2015}%
{Shell \MakeLowercase{\textit{et al.}}: Bare Demo of IEEEtran.cls for Computer Society Journals}

\IEEEtitleabstractindextext{%
\begin{abstract}
 The evolution of convolutional neural networks (CNNs) can be largely attributed to the design of its architecture, i.e., the network wiring pattern. Neural architecture search (NAS) advances this by automating the search for the optimal network architecture, but the resulting network instance may not generalize well in different tasks. To overcome this, exploring network design principles that are generalizable across tasks is a more practical solution. In this study, We explore a novel brain-inspired design principle based on the core-periphery property of the human brain network to guide the design of CNNs. Our work draws inspiration from recent studies suggesting that artificial and biological neural networks may have common principles in optimizing network architecture. We implement the core-periphery principle in the design of network wiring patterns and the sparsification of the convolution operation. The resulting core-periphery principle guided CNNs (CP-CNNs) are evaluated on three different datasets. The experiments demonstrate the effectiveness and superiority compared to CNNs and ViT-based methods. Overall, our work contributes to the growing field of brain-inspired AI by incorporating insights from the human brain into the design of neural networks.
\end{abstract}

}

\maketitle
\IEEEdisplaynontitleabstractindextext
\IEEEpeerreviewmaketitle

\IEEEraisesectionheading{\section{Introduction}
\label{sec:introduction}}

\IEEEPARstart{C}{onvolutional} neural networks (CNNs) have greatly reshaped the paradigm of image processing with impressive performances rivaling human experts in the past decade \cite{lecun1995convolutional,li2014medical,lecun2015deep}. Though with a biologically plausible inspiration from the cat visual cortex\cite{hubel1959receptive,lecun1995convolutional}, the evolution and success of CNNs can be largely attributed to the design of network architecture, i.e., the wiring pattern of neural network and the operation type of network nodes. Early CNNs such as AlexNet \cite{krizhevsky2017imagenet} and VGG \cite{simonyan2014very} adopted a chain-like wiring pattern where the output of the preceding layer is the input of the next layer. Inception CNNs employ an Inception module that concatenates multiple branching pathways with different operations \cite{szegedy2015going,szegedy2016rethinking}. ResNets propose a wiring pattern $x+F(x)$ aiming to learn a residual mapping that enables much deeper networks, and have been widely adapted for many scenarios such as medical imaging with superior performance and generalizability \cite{he2016deep}. Orthogonally, depthwise separable convolution operation greatly reduces the number of parameters and enables extremely deeper CNNs \cite{howard2017mobilenets}. Recent studies also suggest that CNNs can benefit from adopting convolution operation with large kernels (e.g., $7\times7$) \cite{han2021demystifying,liu2022convnet} with comparable performance with Swin Transformer \cite{liu2021swin}. By combining dilated convolution operation and large convolution kernel, a CNN-based architecture can achieve state-of-the-art in some visual tasks \cite{guo2022visual}.

Neural Architecture Search (NAS) advances this trend by jointly optimizing the wiring pattern and the operation to perform. Basically, NAS methods sample from a series of possible architectures and operations through various optimization methods such as reinforcement learning (RL) \cite{zoph2016neural}, evolutionary methods \cite{real2019regularized}, gradient-based methods \cite{liu2018darts}, weight-sharing \cite{pham2018efficient}, and random search \cite{li2020random}. Despite its effectiveness, NAS does not offer a general principle for network architecture design. The outcome of NAS for each run is a neural network instance for a specific task, which may not be generalized to other tasks. For example, an optimal network architecture for natural image classification may not be optimal for X-ray image classification. Hence, some studies explored the design space of neural architectures \cite{radosavovic2020designing} and investigated the general design principles that can be applied to various scenarios. 

Recently, a group of studies suggested that artificial neural networks (ANNs) and biological neural networks (BNNs) may share common principles in optimizing the network architecture. For example, the property of small-world in brain structural and functional networks are recognized and extensively studied in the literature \cite{bassett2006small,bullmore2009complex,bassett2017small}. In \cite{xie2019exploring}, the neural networks based on Watts-Strogatz (WS) random graphs with small-world properties yield competitive performances compared with hand-designed and NAS-optimized models. Through quantitative post-hoc analysis, \cite{you2020graph} found that the graph structure of top-performing ANNs such as CNNs and multilayer perceptron (MLP) is similar to those of real BNNs such as the network in macaque cortex. \cite{zhao2022coupling} synchronized the activation of ANNs and BNNs and found that ANNs with higher performance are similar to BNNs in terms of visual representation activation. Together, these studies suggest the potential of taking advantage of prior knowledge from brain science to guide the architecture design of neural networks.

\begin{figure}[ht]
  \centering
   \includegraphics[width=1.0\linewidth]{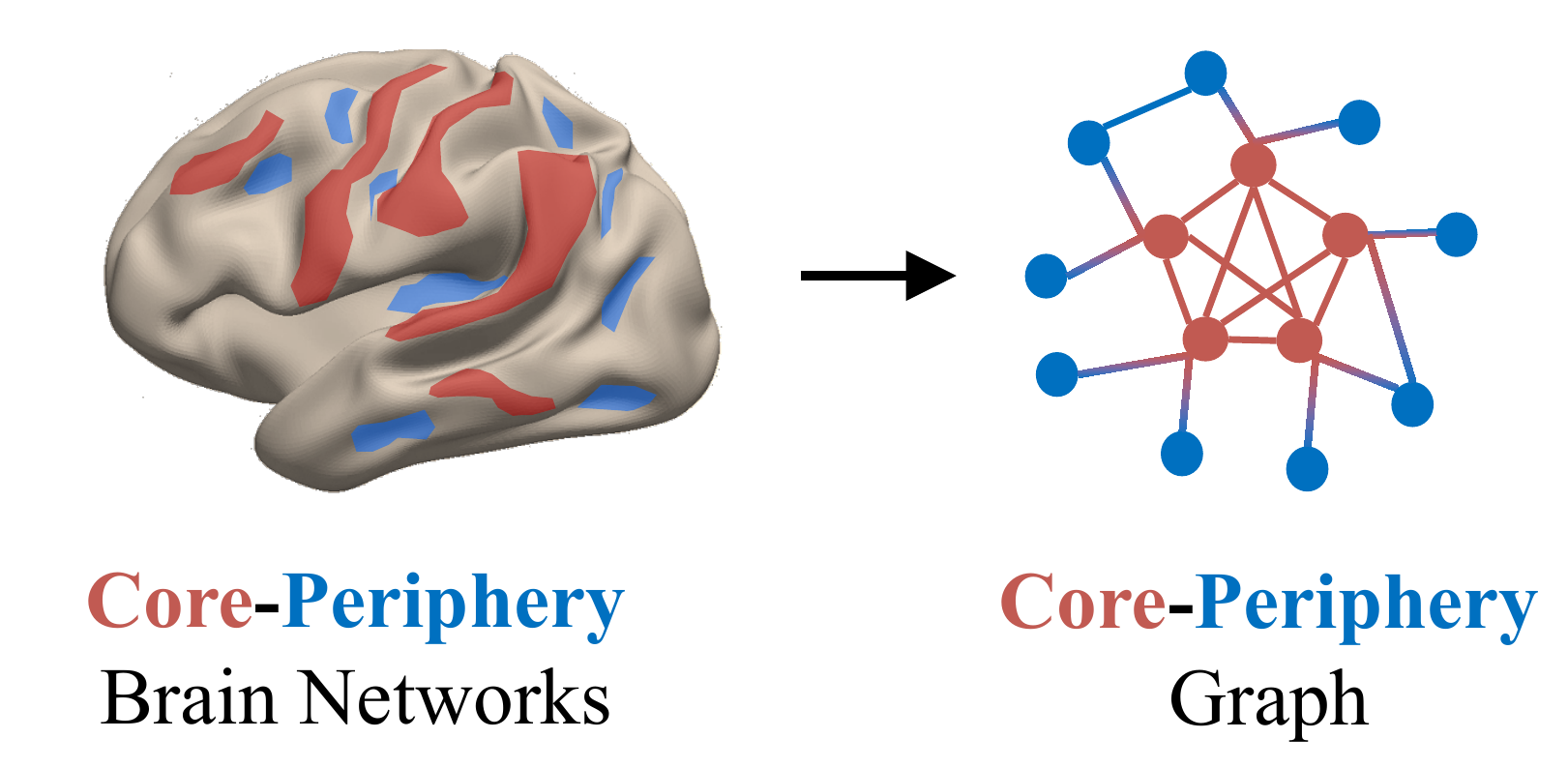}
   \caption{Core-periphery graph. The core nodes are denoted by red color, and the periphery nodes are denoted by blue color.}
   \label{fig:figure1}
\end{figure}

Motivated by these aspects, we explore a brain-inspired Core-Periphery (CP) principle for guiding the architecture design of CNNs. Core-Periphery organization is well-recognized in structural and functional brain networks of humans and other mammals \cite{bassett2013task,gu2020unifying}, which boosts the efficiency of information segregation, transmission and integration. We illustrate the concept of Core-Periphery network organization in \Cref{fig:figure1}. Core-core node pairs have the strongest connection in comparison to core-periphery node pairs (moderate) and periphery-periphery node pairs (the lowest). We design a novel core-periphery graph generator according to this property and introduce a novel core-periphery principle guided CNN (CP-CNN). CP-CNN follows a typical hierarchical scheme of CNNs (e.g., ResNet~\cite{he2016deep}) which consists of a convolutional stem and four consecutive blocks. For each block, we abandon the traditional chain-like wiring pattern but adopt a directed acyclic computational graph which is mapped from the generated core-periphery graph where each node corresponds to an operation such as convolution. In addition, we sparsify the convolution operation in a channel-wise manner and enforce it to follow a core-periphery graph constraint. The proposed CP-CNN is evaluated on CIFAR-10 dataset and two medical imaging datasets (INBreast, NCT-CRC). The experiments demonstrate the effectiveness of the CP-CNN, as well as its superior performance over state-of-the-art CNN and ViT-based methods.

The main contributions of our work are summarized as follows:

\begin{itemize}
\item We proposed a novel brain-inspired CP-CNN model which follows a core-periphery design principle and outperforms the state-of-the-art CNN and ViT baselines.

\item We proposed a core-periphery graph-constrained convolution operation, which reduces the complexity of the model and improves its performances.

\item Our work paves the road for future brain-inspired AI to leverage the prior knowledge from the human brain to inspire the neural network design.

\end{itemize}

\section{Related Works}

\subsection{Neural Architecture of CNNs}
\noindent\textbf{Wiring Pattern.} The development of the wiring pattern significantly contributes to CNN's performance. The early neural architecture of CNNs adopted chain-like wiring patterns, such as AlexNet \cite{krizhevsky2017imagenet} and VGG \cite{simonyan2014very}. Inception \cite{szegedy2015going,szegedy2016rethinking} concatenates several parallel branches with different operations together to "widen" the CNNs. ResNets\cite{he2016deep} propose a wiring pattern $x+F(x)$ for residual learning, which eliminates the gradient vanishing and makes the CNNs much deeper. DenseNet adopted a wiring pattern $[x,F(x)]$ which concatenates the feature maps from the previous layer. The wiring pattern of ResNet and DenseNet is well generalized in various scenarios and applications with improved performances.\newline

\noindent\textbf{Sparsity in Convolution Operation.} Early CNNs used dense connectivity between input and output features, where every output feature is connected to every input feature. To reduce the parameter of such dense connectivity, depthwise separable convolution \cite{howard2017mobilenets} was proposed to decompose the convolution operation as depthwise convolution and 
pointwise convolution, enabling much deeper CNNs. Another group of studies explored the pruning-based method to introduce sparsity in convolution operation, including channel pruning\cite{he2017channel}, filter pruning \cite{luo2018thinet,huang2018learning}, structured pruning \cite{wang2021convolutional}. The introduced sparsity reduced the number of parameters, making the networks easier to train, and also improved their performance on various tasks \cite{hoefler2021sparsity}. \newline

\noindent\textbf{Neural Architecture Search.} NAS jointly optimizes the wiring pattern and the operation to perform. NAS methods predefined a search space, and a series of possible architectures and operations are sampled and selected based on various optimization methods such as reinforcement learning (RL) \cite{zoph2016neural}, evolutionary methods \cite{real2019regularized}, gradient-based methods \cite{liu2018darts}, weight-sharing \cite{pham2018efficient}, and random search \cite{li2020random}. However, the predefined search space still limited the feasible neural architectures to be sampled, regardless of the optimization methods. Meanwhile, the search process usually demands huge computational resources, while the searched architecture may not generalize well for different tasks.

\subsection{Core-Periphery Structure}
Core-periphery structure represents a relationship between nodes in a graph where the core nodes are densely connected with each other while periphery nodes are sparsely connected to the core nodes and among each other \cite{borgatti2000models,rombach2014core}. Core-periphery graph has been applied in a variety of fields, including social network analysis \cite{borgatti2000models,cattani2008core}, economics\cite{kostoska2020core}, biology such as modeling the structure of protein interaction networks\cite{luo2009core}. In the brain science field, it has been shown that brain dynamics has a core-periphery organization \cite{bassett2013task}. The functional brain networks also demonstrate a core-periphery structure\cite{gu2020unifying}. A recent study revealed the core-periphery characteristics of the human brain from a structural perspective \cite{yu2023gyri}. It is shown that gyri and sulci, two prominent cortical folding patterns, could cooperate as a core-periphery network which improves the efficiency of information transmission in the brain \cite{yu2023gyri}.  

\subsection{Connection of ANNs and BNNs}

Recently, a group of studies suggested that artificial neural networks (ANNs) and biological neural networks (BNNs) may share some common principles in optimizing the network architecture. For example,  the property of small-world in brain structural and functional networks are recognized and extensively studied in the literature \cite{bassett2006small,bullmore2009complex,bassett2017small}. Surprisingly, in \cite{xie2019exploring}, the neural networks based on Watts-Strogatz (WS) random graphs with small-world properties yield competitive performances compared with hand-designed and NAS-optimized models. Through quantitative post-hoc analysis, \cite{you2020graph} found that the graph structure of top-performing ANNs such as CNNs and multilayer perceptron (MLP) is similar to those of real BNNs such as the network in macaque cortex. \cite{zhao2022coupling} synchronized the activation of ANNs and BNNs and found that ANNs with higher performance are similar to BNNs in terms of visual representation activation. Together, these studies suggest the potential of taking advantage of prior knowledge from brain science to guide the model architecture design.

\section{Methodology}
\label{method}
In this section, we introduce the generation of the core-periphery graph and present the details of the CP-CNN framework, including the network architecture of CP-CNN, the construction of core-periphery block (CP-Block), and core-periphery constrained convolution operation.

\begin{figure*}[t]
  \centering
   \includegraphics[width=1.0\linewidth]{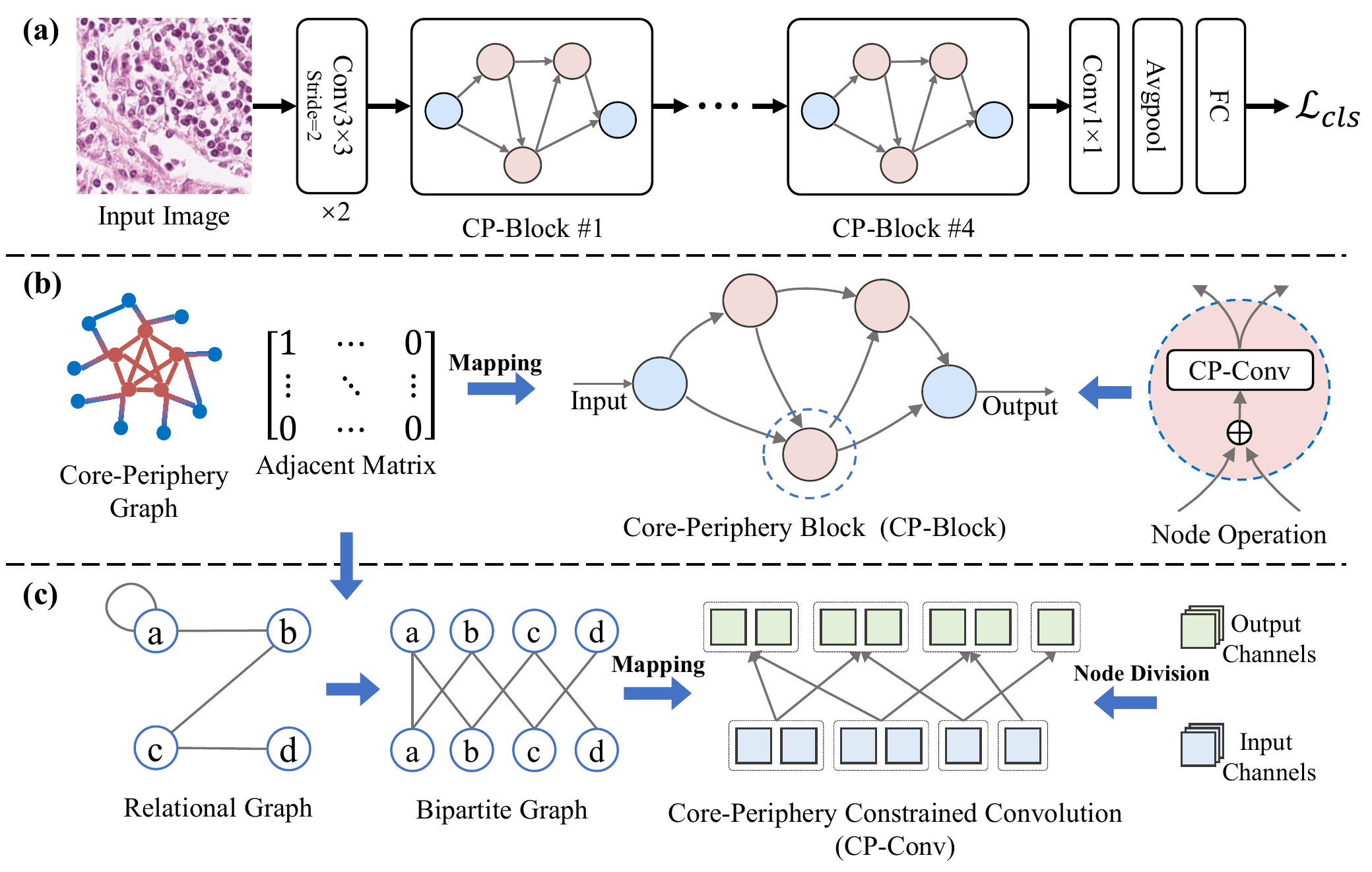}
   \caption{Illustration of the proposed CP-CNN framework. (a) The architecture of the CP-CNN with one convolution stem, four consecutive CP-Blocks, followed by one $1\times1$ convolution, one pooling and one fully-connected layer. (b)The construction of CP-Block and the illustration of the node in CP-Block. The core-periphery graph is mapped as a computational graph for CP-Block based on the node operation. (c) Utilizing core-periphery graph to constrain the convolution operation.}
   \label{fig:figure2}
\end{figure*}

\subsection{Generation of Core-periphery Graph}
\label{sec:generation}
The core-periphery graph (CP graph) has a fundamental signature that the "core-core" node pairs have the strongest interconnections compared with the "core-periphery node" pairs (moderate) and "periphery-periphery" node pairs (weakest). According to this property, we introduce a novel CP graph generator to produce a wide spectrum of CP graphs in this subsection.

Specifically, the proposed CP graph generator is parameterized by the total number of nodes $n$, the number of "core" nodes $n_c$, and the wiring probabilities $p_{cc}$, $p_{cp}$, $p_{pp}$ between "core-core", "core-periphery", "periphery-periphery" node pairs, respectively. The CP graph is generated based on the following process: for each "core-core" node pair, we sample a random number $r$ from a uniform distribution on $[0,1]$. If the wiring probability $p_cc$ is greater than the random number $r$, the "core-core" node pair is connected. The same procedure is also applied to "core-periphery" node pairs and "periphery-periphery" node pairs with the wiring probability $p_{cp}$ and $p_{pp}$, respectively. We summarize the whole generation process in \Cref{alg:cpgraph}. With different combinations of $n$, $n_c$ and wiring probabilities $p_{cc}$, $p_{cp}$, $p_{pp}$, we can generate a wide range of CP graphs in the space, which are then used for constructing the CP-CNN framework introduced in the following subsections.

\begin{algorithm}[h]
\caption{Generation of core-periphery graph}
\label{alg:cpgraph}
\KwIn{$n$: number of nodes; \newline
$n_c$: number of core nodes; \newline
$p_{cc}$, $p_{cp}$, $p_{pp}$: wiring probabilities}
\KwOut{$G$: core-periphery graph}

$G=\emptyset$\;

\tcp{"core-core" node pairs}
\For{$i\gets0$ \KwTo $n_c$}{
    \For{$j \gets i$ \KwTo $n_c$}{
    Sample a uniform random number $r\in[0,1)$
    \If{$r<p_{cc}$}{
    $G\gets (i,j)$
    }
  }
}

\tcp{"core-periphery" node pairs}
\For{$i\gets0$ \KwTo $n_c$}{
    \For{$j \gets n_c$ \KwTo $n$}{
    Sample a uniform random number $r\in[0,1)$
    \If{$r<p_{cp}$}{
    $G\gets (i,j)$
    }
  }
}
\tcp{"periphery-periphery" node pairs}
\For{$i\gets n_c$ \KwTo $n$}{
    \For{$j \gets i$ \KwTo $n$}{
    Sample a uniform random number $r\in[0,1)$
    \If{$r<p_{pp}$}{
    $G\gets (i,j)$
    }
  }
}

\Return{$G$}

\end{algorithm}

\subsection{CP-CNN Framework}
\label{sec:cpcnn}
Our macro design of CP-CNN architecture follows a typical hierarchical scheme of CNNs (e.g., ResNet~\cite{he2016deep}) with a convolutional stem and several convolution blocks (\Cref{fig:figure2}(a)). Specifically, the input image is firstly input into a convolution stem which consists of two $3\times 3$ convolutions with a stride of 2. The feature maps from the convolution stem are then processed by four consecutive core-periphery blocks (CP-Blocks, discussed in detail in \Cref{sec:cpblock} below). Within each CP-Block, the size of the feature map is decreased by $2\times$ while the number of channels is increased by $2\times$. A classification head with $1\times1$ convolution, global average pooling and a fully connected layer is added after the CP-Block to produce the final prediction. 

\subsection{Core-periphery Block}
\label{sec:cpblock}
Unlike the traditional chain-like structure, our core-periphery block has a "graph" structure (\Cref{fig:figure2}(b)) which is implemented based on the generated core-periphery graph. To construct the core-periphery block, we need to convert the generated core-periphery graph into computational graph in the neural network. However, the generated core-periphery graphs are undirected while the computational graph in neural networks are directed and acyclic. So the first step is to convert the generated core-periphery graph into a directed acyclic graph (DAG), and then map the DAG into a computation graph for the CP-Block.

Specifically, we adopt a heuristic strategy to perform such conversion. For each node in the core-periphery graph, we randomly assign a unique label ranging from 1 to $n$ (the number of nodes in the graph) to it. Then, for all undirected edges in the graph, we convert it into directed edges which always start from the node with the small label and end with the node having the large label. This approach guarantees that there are no circles in the resulting directed graph, i.e., the resulting graph is a DAG. The next step is to map the DAG into a computational graph in the neural network. To do so, we first need to define the node and edge in the computational graph. 

\noindent\textbf{Edges.} Similar to edges in most computation graphs, we define that the directed edge in our implementation represents the direction of data flow, i.e., the node sends the data to another node along this flow.

\noindent\textbf{Nodes.} We define the nodes in our computational graph as processing units that aggregate and process the data from input edges and distribute the processed data to other nodes along the output edges. As illustrated in \Cref{fig:figure2}(c), the data tensors along the input edges are firstly aggregated through a weighted sum. The weights of the aggregation are learnable. Then, the combined tensors are processed by an operation unit which consists of ReLU activation, $3\times3$ core-periphery convolution (discussed in detail in \Cref{sec:cpconv} below), and batch normalization. The unit's output is distributed as the same copies to other nodes along the output edges.

Using the defined nodes and edges, we obtain an intermediate computational graph. However, this graph may have several input nodes (those without input edges) and output nodes (those without output edges), while each block is expected to have only one input and one output. To address this, we introduce an additional input node that performs convolution with a stride of 2 on the previous block's output or the convolution stem, sending the same feature maps to all original input nodes. Similarly, we introduce an output node that aggregates the feature maps from all original output nodes using a learnable weighted sum, without performing any convolution within this node. This creates the CP-Block, which can be stacked in the CP-CNN as previously discussed.

\subsection{Core-periphery Constrained Convolution}
\label{sec:cpconv}

The CP-Block can also be constructed using conventional convolution in the nodes of the computational graph. However, traditional convolution is more "dense" whereas incorporating sparsity into the neural network can significantly lower its complexity and enhance its performance, especially in scenarios with limited training samples such as medical imaging.

Inspired by this, we propose a novel Core-Periphery Constrained Convolution that utilizes a core-periphery graph as a constraint to sparsify the convolution operation. Specifically, we divide the input and output channels of the convolution into $n$ groups and represent the relationship between them as a bipartite graph (\Cref{fig:figure2}(c)). In conventional convolution, the bipartite graph is densely connected, with all input channels in a filter contributing to the production of all output channels. For example, output channels in node $\#1$ integrate information from all input channels. In contrast, a sparse bipartite graph means that only a portion of input channels is used to generate output channels. As shown in \Cref{fig:figure2}(c), the output channels in node $\#1$ only integrate information from input channels in node $\#1$ and node $\#2$. By sparsifying the convolution operation with a predefined bipartite graph, the convolution is constrained by a graph.S

We use the core-periphery graph as a constraint by converting the generated graph into a bipartite graph. The core-periphery graph is first represented as the relational graph proposed in \cite{you2020graph} which represents the message passing between nodes. The relational graph is then transformed into a bipartite graph, where the nodes in two sets correspond to the divided sets of input and output channels, respectively. The edges in the bipartite graph represent message passing in the relational graph. We apply the resulting bipartite graph as a constraint to the convolution operation to obtain the core-periphery constrained convolution. It is worth noting that we apply the same core-periphery graph across the whole network, while the constrained convolution may vary among different nodes and blocks due to the varying number of channels.

\section{Experiments}

\textbf{Datasets.}
We evaluate the proposed framework on three datasets, including one for natural images and two for medical images. \textbf{CIFAR-10} \cite{krizhevsky2009learning} consists of 60,000 $32\times32$ images in 10 classes, with 50,000 images in the training set and 10,000 images in the test set. In our experiments, we upsample all original images in CIFAR-10 to $224\times224$. \textbf{NCT-CRC} \cite{kather2018100} contains 100,000 non-overlapping training image patches extracted from hematoxylin and eosin (H\&E) stained histological images of human colorectal cancer (CRC) and normal tissue \cite{kather2018100}. Additional 7,180 image patches from 50 patients with no overlap with patients in the training set are used as a validation set. Both training and validation sets have 9 classes and size of $224\times224$ for each patch. \textbf{INbreast} dataset~\cite{moreira2012inbreast} includes 410 full-ﬁeld digital mammography images collected during low-dose X-ray irradiation of the breast. These images can be classified into normal (302 cases), benign (37 cases), and malignant (71 cases) classes. We randomly split the patients into 80\% and 20\% as training and testing datasets. To balance the training dataset, we perform several random cropping with a size of $1024\times1024$ as well as the contrast-related augmentation for each image, resulting in 482 normal samples, 512 benign mass samples, and 472 malignant mass samples. The images in both sets are downsized into $224\times224$.\newline

\noindent\textbf{Implementation Details.}
In our experiments, we set the number of nodes in the core-periphery graph to 16 and vary the number of core nodes. The three probabilities are set as $p_{cc}=0.9$, $p_{cp}=0.5$, $p_{cc}=0.1$. The proposed model and all compared baselines are trained for 50 epochs with a batch size of 512. We use the AdamW optimizer~\cite{kingma2014adam} with $\beta_1=0.9$ and $\beta_2=0.999$ and a cosine annealing learning rate scheduler with initial learning $10^{-4}$ and 5 warm-up epochs. The framework is implemented with PyTorch (\url{https://pytorch.org/}) deep learning library and the model is trained on 4 NVIDIA A5000 GPU. 

\subsection{Comparison with Baselines}
To validate the proposed CP-CNN, we compare the performance of CP-CNN with various state-of-the-art baselines, which can be roughly categorized as CNN-based methods and ViT-based methods. CNN-based category contains ResNet\cite{he2016deep}, EfficientNet~\cite{tan2019efficientnet}, RegNet\cite{radosavovic2020designing}, ConvNeXt\cite{liu2022convnet}. The ViT-based class contains vanilla ViT\cite{dosovitskiy2020image}, CaiT\cite{touvron2021going} and Swin Transformer\cite{liu2021swin}. Considering the amount of data, we set the number of nodes in the CP graph as 16, resulting CP-CNN model with around 22 million parameters. For the compared methods, we re-implement them and only report the tiny- or small-scale setting with comparable parameters as CP-CNN.

\Cref{tab:tab1} demonstrates a comprehensive comparison of the Top-1 classification accuracy (\%) achieved by different models on three datasets, as well as the number of parameters and flops. It is observed that CNNs generally exhibit superior performance compared to ViTs. This can be attributed to the inductive biases in CNNs, which are essential in scenarios with a limited number of training samples. This is also suggested by the observation that SwinV2-T, which incorporates inductive biases, outperforms other ViT models. 

Our proposed CP-CNN model achieves state-of-the-art performance compared to other CNN-based methods, demonstrating its superiority in terms of accuracy. Specifically, our CP-CNN outperforms baseline models in all settings on the CIFAR-10 dataset. For the NCT-CRC dataset, our CP-CNN model achieves higher accuracy compared to both CNNs and ViTs, except for sparse settings with only 2 or 4 core nodes. Furthermore, on the INBreast dataset, our sparse CP-CNN model with 2 core nodes achieves state-of-the-art performance. Importantly, our CP-CNN model's superior performance is achieved while requiring a comparable number of parameters and flops as other models. Thus, the proposed CP-CNN can be a promising solution for image classification tasks, offering both high accuracy and efficiency. 

It is also noted that our CP-CNN model outperforms the RegNet model which is also based on the exploration of design space of neural architecture. This indicates that the brain-inspired core-periphery design principle may be more generalized than the empirical design principles as those in RegNet.

\begin{table*}[ht]
\centering
\caption{Top-1 classification accuracy (\%) of proposed and compared models on the CIFAR-10, NCT-CRC, and INBreast datasets, along with the number of parameters (M) and flops (G). The models with the highest accuracy are highlighted in \textbf{bold}. For some settings, the models do not get converged and are indicated by a slash (/) symbol.}
\label{tab:tab1}
\begin{tabular}{clccccc}
\toprule

Category & Models & CIFAR-10\cite{krizhevsky2009learning} & NCT-CRC\cite{kather2018100} & INBreast\cite{moreira2012inbreast} & Param (M) & Flops (G)\\
\midrule

\multirow{8}{*}{CNNs} 
& ResNet-18\cite{he2016deep} &90.35 &95.96 &83.56 &11.18&1.8\\
& ResNet-50\cite{he2016deep} &90.55 &95.11 &82.19 &23.53&4.1\\
& EfficientNet-B3\cite{tan2019efficientnet} &82.52 &95.25&/&10.71&1.8\\
& EfficientNet-B4\cite{tan2019efficientnet} &81.73 &95.21&/&17.56&4.4\\
& RegNetY-016 &88.03 &95.91 &/&10.32&1.6\\
& RegNetX-032 &88.78 &95.91 &/&14.30&3.2\\
& ConvNeXt-Nano\cite{liu2022convnet} &86.88 &95.10 &/&14.96&2.5\\
& ConvNeXt-Tiny\cite{liu2022convnet} &86.32 &94.64 &/&27.83&4.5\\

\midrule

\multirow{5}{*}{ViTs} 
& ViT-Tiny\cite{dosovitskiy2020image} &76.10&90.63 &/&5.50&1.3\\
& ViT-Small\cite{dosovitskiy2020image} &69.37&89.79&/&21.67&4.6\\
& CaiT-XXS-24\cite{touvron2021going} &73.99 &92.06&/&11.77&2.5\\
& CaiT-XXS-36\cite{touvron2021going} &74.36 &92.41&/&17.11&3.8\\
& SwinV2-T\cite{liu2022swin}&81.76 &95.61 &/&27.57&5.9\\
\midrule

\multirow{7}{*}{CP-CNN} 
& N=16, C=2  &91.22 &95.28 &\textbf{85.75} &22.21&3.4\\
& N=16, C=4  &91.71 &95.43 &82.19 &22.21&3.4\\
& N=16, C=6  &91.99 &96.34 &83.01 &22.21&3.4\\
& N=16, C=8  &92.41 &\textbf{96.78} &83.01 &22.21&3.4\\
& N=16, C=10 &94.43 &96.65 &83.28 &22.21&3.4\\
& N=16, C=12 &92.54 &96.29 &83.56 &22.21&3.4\\
& N=16, C=14 &\textbf{92.65} &96.60 &84.11 &22.21&3.4\\

\bottomrule
\end{tabular}
\end{table*}

\subsection{Sparsity of CP Graph}
The number of core nodes $c$ controls the sparsity of the generated CP graph. More core nodes indicate the dense connections in the CP graph. In this subsection, we investigate the effects of CP graph sparsity on classification performances.

As illustrated in \Cref{tab:tab1}, we fix the number of total nodes to 16 and vary the number of core nodes from 2 to 14 in steps of 2, resulting in graph sparsity ranging from 0.125 to 0.875 with an interval of 0.125. For the CIFAR-10 dataset, we observed an increase in classification accuracy with the increase in the number of core nodes, reaching a peak with 14 core nodes (sparsity=0.875). It is probably because a dense graph increases the capacity of the CP-CNN model, so it can represent more complex relationships. In contrast, for the INBreast dataset, the sparsest CP-CNN model (2 core nodes, sparsity=0.125) yields the best performances. This may be due to the dataset having only thousands of training samples. A large and dense model may suffer from the overfitting problem, which reduces performance. For the NCT-CRC dataset, the performance increased with sparsity, with the highest accuracy achieved at a sparsity of 0.5. The accuracy slightly decreased with a more dense graph. This may be because a sparse model with low capacity may not be able to represent the complex relationships in the dataset, while a dense model may overfit. At a sparsity of 0.5, the right balance between model capacity and dataset complexity was achieved. Overall, the sparsity of the CP Graph can affect the capacity of the CP-CNN model and, thus, the performances on different datasets. Despite this, the CP-CNN model still has comparable and superior performances compared with baseline models.

\subsection{Comparison with Random Graphs}
\begin{figure}[ht]
  \centering
   \includegraphics[width=1.0\linewidth]{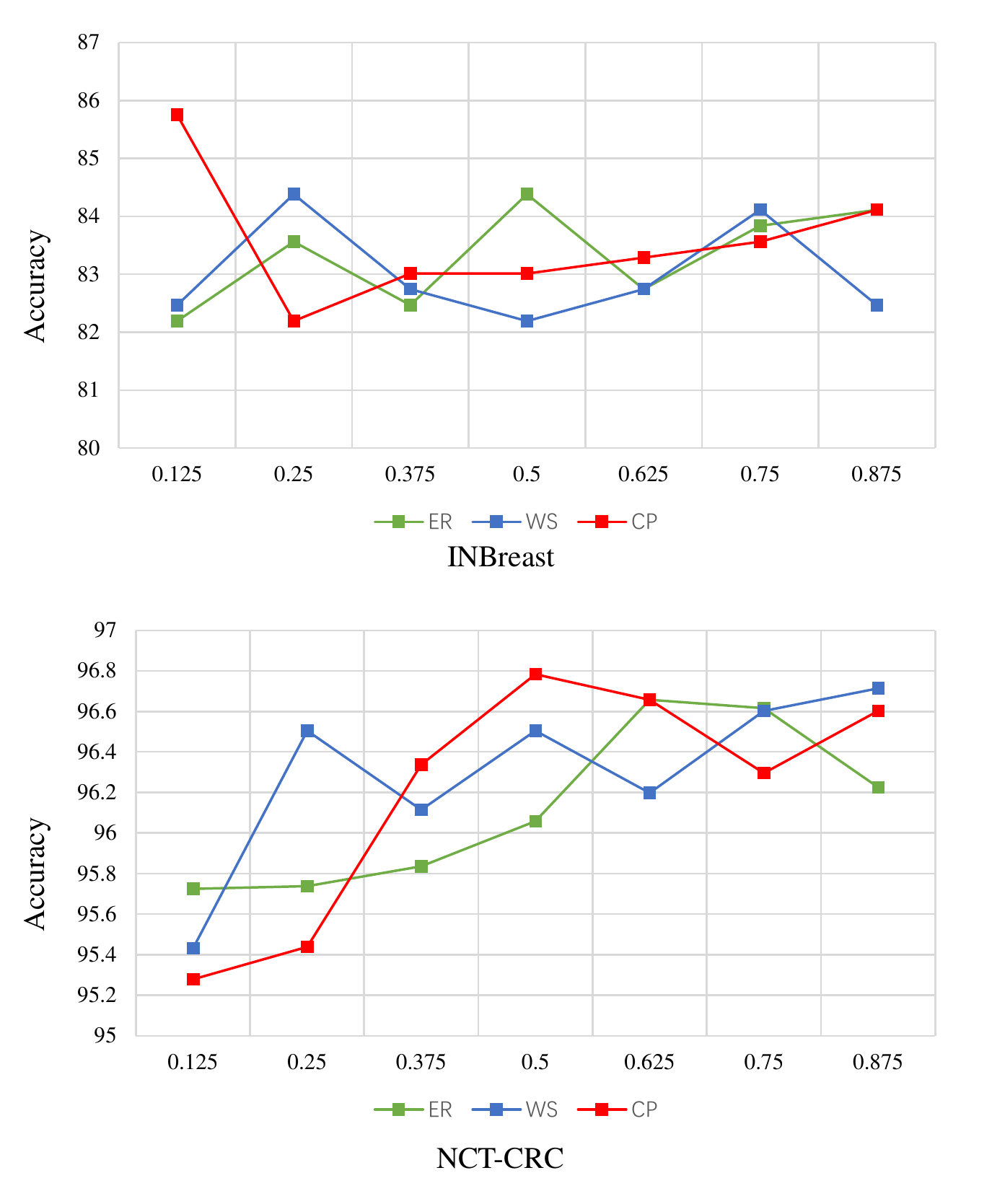}
   \caption{The comparison of ER, WS, and CP graph with varying sparsity based on the CP-CNN model, in terms of accuracy, using the INBreast and NCT-CRC datasets.}
   \label{fig:figure3}
\end{figure}

To validate the effectiveness of the CP graph, we replace the CP graph in the CP-CNN model with two random graphs: Erdos-Renyi (ER) graph \cite{erdos1960evolution} and Watts-Strogatz (WS) graph\cite{watts1998collective}. ER graph is parameterized by $P$, which is the probability that two nodes are connected by an edge. WS graph is considered to have the small-world property. In our experiment, we randomly generate 10 samples for ER, WS, and CP graphs with the same sparsity. In \Cref{fig:figure3}, we demonstrate the average classification accuracy across the 10 samples for different graphs and sparsity.

It is observed that the CP graph with a sparsity of 0.125 outperforms all other settings and graphs on the INBreast dataset, whereas on other sparsity settings, different graphs achieve the best accuracy. For the NCT-CRC dataset, the CP graph outperforms the ER and WS graphs with sparsity values of 0.375, 0.5, and 0.625, and achieves the highest accuracy among all settings and graphs with a sparsity of 0.5. These results suggest that the choice of graph type and sparsity can significantly affect the performance of the CP-CNN model on different datasets. However, with specific sparsity settings, CP graph can provide superior performance compared to ER and WS graphs, i.e., the CP graph has an upper performance bound than ER and WS graphs.

In addition, the CP-CNN models based on ER and WS graphs also have competitive performances than the CNNs and ViTs in \Cref{tab:tab1}], highlighting the potential of incorporating graph structures in CNNs for improving their performance and generalization ability.

\section{Discussion}

\textbf{Brain-inspired AI.} The brain is a highly complex network of interconnected neurons that communicate with each other to process and transmit information. Core-periphery property is a representative signature of the brain network. The results reported in the study suggest that incorporating the properties/principles of brain networks can effectively improve the performance of CNNs. Our study provides a promising solution and contributes to brain-inspired AI by leveraging the prior knowledge of the human brain to inspire the design of ANNs.\newline

\noindent\textbf{Limitations.}
The sparsity of the CP graph can affect the capacity of CP-CNN model. The experiments demonstrated that the optimal capacity of the CP-CNN model may vary depending on the dataset and the specific problem being solved. Line and grid search may help us to determine the optimal sparsity for different datasets. However, how to effectively search the optimal sparsity is still an opening question. In addition, the proposed CP-CNN model is evaluated at a scale of 22 million parameters, which is suitable for relatively small datasets, especially those in medical imaging scenarios. The performances of a larger-scale CP-CNN model on a larger dataset, such as ImageNet-1K will be investigated in the future.

\section{Conclusion}

In this study, we explored a novel brain-inspired core-periphery design principle to guide the design of CNNs. The core-periphery principle was implemented in both the design of network wiring patterns and the sparsification of the convolution operation. The experiments demonstrate the effectiveness and superiority of the CP principle-guided CNNs compared to CNNs and ViT-based methods. In general, our study advances the growing field of brain-inspired artificial intelligence by integrating prior knowledge from the human brain to inspire the design of artificial neural networks.

% References
\bibliographystyle{IEEEtran}
\bibliography{refs}

\begin{thebibliography}{10}
\providecommand{\url}[1]{#1}
\csname url@samestyle\endcsname
\providecommand{\newblock}{\relax}
\providecommand{\bibinfo}[2]{#2}
\providecommand{\BIBentrySTDinterwordspacing}{\spaceskip=0pt\relax}
\providecommand{\BIBentryALTinterwordstretchfactor}{4}
\providecommand{\BIBentryALTinterwordspacing}{\spaceskip=\fontdimen2\font plus
\BIBentryALTinterwordstretchfactor\fontdimen3\font minus
  \fontdimen4\font\relax}
\providecommand{\BIBforeignlanguage}[2]{{%
\expandafter\ifx\csname l@#1\endcsname\relax
\typeout{** WARNING: IEEEtran.bst: No hyphenation pattern has been}%
\typeout{** loaded for the language `#1'. Using the pattern for}%
\typeout{** the default language instead.}%
\else
\language=\csname l@#1\endcsname
\fi
#2}}
\providecommand{\BIBdecl}{\relax}
\BIBdecl

\bibitem{lecun1995convolutional}
Y.~LeCun, Y.~Bengio \emph{et~al.}, ``Convolutional networks for images, speech,
  and time series,'' \emph{The handbook of brain theory and neural networks},
  vol. 3361, no.~10, p. 1995, 1995.

\bibitem{li2014medical}
Q.~Li, W.~Cai, X.~Wang, Y.~Zhou, D.~D. Feng, and M.~Chen, ``Medical image
  classification with convolutional neural network,'' in \emph{2014 13th
  international conference on control automation robotics \& vision
  (ICARCV)}.\hskip 1em plus 0.5em minus 0.4em\relax IEEE, 2014, pp. 844--848.

\bibitem{lecun2015deep}
Y.~LeCun, Y.~Bengio, and G.~Hinton, ``Deep learning,'' \emph{nature}, vol. 521,
  no. 7553, pp. 436--444, 2015.

\bibitem{hubel1959receptive}
D.~H. Hubel and T.~N. Wiesel, ``Receptive fields of single neurones in the
  cat's striate cortex,'' \emph{The Journal of physiology}, vol. 148, no.~3, p.
  574, 1959.

\bibitem{krizhevsky2017imagenet}
A.~Krizhevsky, I.~Sutskever, and G.~E. Hinton, ``Imagenet classification with
  deep convolutional neural networks,'' \emph{Communications of the ACM},
  vol.~60, no.~6, pp. 84--90, 2017.

\bibitem{simonyan2014very}
K.~Simonyan and A.~Zisserman, ``Very deep convolutional networks for
  large-scale image recognition,'' \emph{arXiv preprint arXiv:1409.1556}, 2014.

\bibitem{szegedy2015going}
C.~Szegedy, W.~Liu, Y.~Jia, P.~Sermanet, S.~Reed, D.~Anguelov, D.~Erhan,
  V.~Vanhoucke, and A.~Rabinovich, ``Going deeper with convolutions,'' in
  \emph{Proceedings of the IEEE conference on computer vision and pattern
  recognition}, 2015, pp. 1--9.

\bibitem{szegedy2016rethinking}
C.~Szegedy, V.~Vanhoucke, S.~Ioffe, J.~Shlens, and Z.~Wojna, ``Rethinking the
  inception architecture for computer vision,'' in \emph{Proceedings of the
  IEEE conference on computer vision and pattern recognition}, 2016, pp.
  2818--2826.

\bibitem{he2016deep}
K.~He, X.~Zhang, S.~Ren, and J.~Sun, ``Deep residual learning for image
  recognition,'' in \emph{Proceedings of the IEEE conference on computer vision
  and pattern recognition}, 2016, pp. 770--778.

\bibitem{howard2017mobilenets}
A.~G. Howard, M.~Zhu, B.~Chen, D.~Kalenichenko, W.~Wang, T.~Weyand,
  M.~Andreetto, and H.~Adam, ``Mobilenets: Efficient convolutional neural
  networks for mobile vision applications,'' \emph{arXiv preprint
  arXiv:1704.04861}, 2017.

\bibitem{han2021demystifying}
Q.~Han, Z.~Fan, Q.~Dai, L.~Sun, M.-M. Cheng, J.~Liu, and J.~Wang,
  ``Demystifying local vision transformer: Sparse connectivity, weight sharing,
  and dynamic weight,'' \emph{arXiv preprint arXiv:2106.04263}, 2021.

\bibitem{liu2022convnet}
Z.~Liu, H.~Mao, C.-Y. Wu, C.~Feichtenhofer, T.~Darrell, and S.~Xie, ``A convnet
  for the 2020s,'' in \emph{Proceedings of the IEEE/CVF Conference on Computer
  Vision and Pattern Recognition}, 2022, pp. 11\,976--11\,986.

\bibitem{liu2021swin}
Z.~Liu, Y.~Lin, Y.~Cao, H.~Hu, Y.~Wei, Z.~Zhang, S.~Lin, and B.~Guo, ``Swin
  transformer: Hierarchical vision transformer using shifted windows,'' in
  \emph{Proceedings of the IEEE/CVF International Conference on Computer
  Vision}, 2021, pp. 10\,012--10\,022.

\bibitem{guo2022visual}
M.-H. Guo, C.-Z. Lu, Z.-N. Liu, M.-M. Cheng, and S.-M. Hu, ``Visual attention
  network,'' \emph{arXiv preprint arXiv:2202.09741}, 2022.

\bibitem{zoph2016neural}
B.~Zoph and Q.~V. Le, ``Neural architecture search with reinforcement
  learning,'' \emph{arXiv preprint arXiv:1611.01578}, 2016.

\bibitem{real2019regularized}
E.~Real, A.~Aggarwal, Y.~Huang, and Q.~V. Le, ``Regularized evolution for image
  classifier architecture search,'' in \emph{Proceedings of the aaai conference
  on artificial intelligence}, vol.~33, no.~01, 2019, pp. 4780--4789.

\bibitem{liu2018darts}
H.~Liu, K.~Simonyan, and Y.~Yang, ``Darts: Differentiable architecture
  search,'' \emph{arXiv preprint arXiv:1806.09055}, 2018.

\bibitem{pham2018efficient}
H.~Pham, M.~Guan, B.~Zoph, Q.~Le, and J.~Dean, ``Efficient neural architecture
  search via parameters sharing,'' in \emph{International conference on machine
  learning}.\hskip 1em plus 0.5em minus 0.4em\relax PMLR, 2018, pp. 4095--4104.

\bibitem{li2020random}
L.~Li and A.~Talwalkar, ``Random search and reproducibility for neural
  architecture search,'' in \emph{Uncertainty in artificial
  intelligence}.\hskip 1em plus 0.5em minus 0.4em\relax PMLR, 2020, pp.
  367--377.

\bibitem{radosavovic2020designing}
I.~Radosavovic, R.~P. Kosaraju, R.~Girshick, K.~He, and P.~Doll{\'a}r,
  ``Designing network design spaces,'' in \emph{Proceedings of the IEEE/CVF
  conference on computer vision and pattern recognition}, 2020, pp.
  10\,428--10\,436.

\bibitem{bassett2006small}
D.~S. Bassett and E.~Bullmore, ``Small-world brain networks,'' \emph{The
  neuroscientist}, vol.~12, no.~6, pp. 512--523, 2006.

\bibitem{bullmore2009complex}
E.~Bullmore and O.~Sporns, ``Complex brain networks: graph theoretical analysis
  of structural and functional systems,'' \emph{Nature reviews neuroscience},
  vol.~10, no.~3, pp. 186--198, 2009.

\bibitem{bassett2017small}
D.~S. Bassett and E.~T. Bullmore, ``Small-world brain networks revisited,''
  \emph{The Neuroscientist}, vol.~23, no.~5, pp. 499--516, 2017.

\bibitem{xie2019exploring}
S.~Xie, A.~Kirillov, R.~Girshick, and K.~He, ``Exploring randomly wired neural
  networks for image recognition,'' in \emph{Proceedings of the IEEE/CVF
  International Conference on Computer Vision}, 2019, pp. 1284--1293.

\bibitem{you2020graph}
J.~You, J.~Leskovec, K.~He, and S.~Xie, ``Graph structure of neural networks,''
  in \emph{International Conference on Machine Learning}.\hskip 1em plus 0.5em
  minus 0.4em\relax PMLR, 2020, pp. 10\,881--10\,891.

\bibitem{zhao2022coupling}
L.~Zhao, H.~Dai, Z.~Wu, Z.~Xiao, L.~Zhang, D.~W. Liu, X.~Hu, X.~Jiang, S.~Li,
  D.~Zhu \emph{et~al.}, ``Coupling visual semantics of artificial neural
  networks and human brain function via synchronized activations,'' \emph{arXiv
  preprint arXiv:2206.10821}, 2022.

\bibitem{bassett2013task}
D.~S. Bassett, N.~F. Wymbs, M.~P. Rombach, M.~A. Porter, P.~J. Mucha, and S.~T.
  Grafton, ``Task-based core-periphery organization of human brain dynamics,''
  \emph{PLoS computational biology}, vol.~9, no.~9, p. e1003171, 2013.

\bibitem{gu2020unifying}
S.~Gu, C.~H. Xia, R.~Ciric, T.~M. Moore, R.~C. Gur, R.~E. Gur, T.~D.
  Satterthwaite, and D.~S. Bassett, ``Unifying the notions of modularity and
  core--periphery structure in functional brain networks during youth,''
  \emph{Cerebral Cortex}, vol.~30, no.~3, pp. 1087--1102, 2020.

\bibitem{he2017channel}
Y.~He, X.~Zhang, and J.~Sun, ``Channel pruning for accelerating very deep
  neural networks,'' in \emph{Proceedings of the IEEE international conference
  on computer vision}, 2017, pp. 1389--1397.

\bibitem{luo2018thinet}
J.-H. Luo, H.~Zhang, H.-Y. Zhou, C.-W. Xie, J.~Wu, and W.~Lin, ``Thinet:
  pruning cnn filters for a thinner net,'' \emph{IEEE transactions on pattern
  analysis and machine intelligence}, vol.~41, no.~10, pp. 2525--2538, 2018.

\bibitem{huang2018learning}
Q.~Huang, K.~Zhou, S.~You, and U.~Neumann, ``Learning to prune filters in
  convolutional neural networks,'' in \emph{2018 IEEE Winter Conference on
  Applications of Computer Vision (WACV)}.\hskip 1em plus 0.5em minus
  0.4em\relax IEEE, 2018, pp. 709--718.

\bibitem{wang2021convolutional}
Z.~Wang, C.~Li, and X.~Wang, ``Convolutional neural network pruning with
  structural redundancy reduction,'' in \emph{Proceedings of the IEEE/CVF
  Conference on Computer Vision and Pattern Recognition}, 2021, pp.
  14\,913--14\,922.

\bibitem{hoefler2021sparsity}
T.~Hoefler, D.~Alistarh, T.~Ben-Nun, N.~Dryden, and A.~Peste, ``Sparsity in
  deep learning: Pruning and growth for efficient inference and training in
  neural networks,'' \emph{The Journal of Machine Learning Research}, vol.~22,
  no.~1, pp. 10\,882--11\,005, 2021.

\bibitem{borgatti2000models}
S.~P. Borgatti and M.~G. Everett, ``Models of core/periphery structures,''
  \emph{Social networks}, vol.~21, no.~4, pp. 375--395, 2000.

\bibitem{rombach2014core}
M.~P. Rombach, M.~A. Porter, J.~H. Fowler, and P.~J. Mucha, ``Core-periphery
  structure in networks,'' \emph{SIAM Journal on Applied mathematics}, vol.~74,
  no.~1, pp. 167--190, 2014.

\bibitem{cattani2008core}
G.~Cattani and S.~Ferriani, ``A core/periphery perspective on individual
  creative performance: Social networks and cinematic achievements in the
  hollywood film industry,'' \emph{Organization science}, vol.~19, no.~6, pp.
  824--844, 2008.

\bibitem{kostoska2020core}
O.~Kostoska, S.~Mitikj, P.~Jovanovski, and L.~Kocarev, ``Core-periphery
  structure in sectoral international trade networks: A new approach to an old
  theory,'' \emph{PloS one}, vol.~15, no.~4, p. e0229547, 2020.

\bibitem{luo2009core}
F.~Luo, B.~Li, X.-F. Wan, and R.~H. Scheuermann, ``Core and periphery
  structures in protein interaction networks,'' in \emph{BMC bioinformatics},
  vol.~10, no.~4.\hskip 1em plus 0.5em minus 0.4em\relax BioMed Central, 2009,
  pp. 1--11.

\bibitem{yu2023gyri}
X.~Yu, L.~Zhang, H.~Dai, L.~Zhao, Y.~Lyu, Z.~Wu, T.~Liu, and D.~Zhu, ``Gyri vs.
  sulci: Disentangling brain core-periphery functional networks via
  twin-transformer,'' \emph{arXiv preprint arXiv:2302.00146}, 2023.

\bibitem{krizhevsky2009learning}
A.~Krizhevsky, G.~Hinton \emph{et~al.}, ``Learning multiple layers of features
  from tiny images,'' 2009.

\bibitem{kather2018100}
J.~N. Kather, N.~Halama, and A.~Marx, ``100,000 histological images of human
  colorectal cancer and healthy tissue,'' \emph{Zenodo10}, vol. 5281, 2018.

\bibitem{moreira2012inbreast}
I.~C. Moreira, I.~Amaral, I.~Domingues, A.~Cardoso, M.~J. Cardoso, and J.~S.
  Cardoso, ``Inbreast: toward a full-field digital mammographic database,''
  \emph{Academic radiology}, vol.~19, no.~2, pp. 236--248, 2012.

\bibitem{kingma2014adam}
D.~P. Kingma and J.~Ba, ``Adam: A method for stochastic optimization,''
  \emph{arXiv preprint arXiv:1412.6980}, 2014.

\bibitem{tan2019efficientnet}
M.~Tan and Q.~Le, ``Efficientnet: Rethinking model scaling for convolutional
  neural networks,'' in \emph{International conference on machine
  learning}.\hskip 1em plus 0.5em minus 0.4em\relax PMLR, 2019, pp. 6105--6114.

\bibitem{dosovitskiy2020image}
A.~Dosovitskiy, L.~Beyer, A.~Kolesnikov, D.~Weissenborn, X.~Zhai,
  T.~Unterthiner, M.~Dehghani, M.~Minderer, G.~Heigold, S.~Gelly \emph{et~al.},
  ``An image is worth 16x16 words: Transformers for image recognition at
  scale,'' \emph{arXiv preprint arXiv:2010.11929}, 2020.

\bibitem{touvron2021going}
H.~Touvron, M.~Cord, A.~Sablayrolles, G.~Synnaeve, and H.~J{\'e}gou, ``Going
  deeper with image transformers,'' in \emph{Proceedings of the IEEE/CVF
  International Conference on Computer Vision}, 2021, pp. 32--42.

\bibitem{liu2022swin}
Z.~Liu, H.~Hu, Y.~Lin, Z.~Yao, Z.~Xie, Y.~Wei, J.~Ning, Y.~Cao, Z.~Zhang,
  L.~Dong \emph{et~al.}, ``Swin transformer v2: Scaling up capacity and
  resolution,'' in \emph{Proceedings of the IEEE/CVF conference on computer
  vision and pattern recognition}, 2022, pp. 12\,009--12\,019.

\bibitem{erdos1960evolution}
P.~Erdos, A.~R{\'e}nyi \emph{et~al.}, ``On the evolution of random graphs,''
  \emph{Publ. Math. Inst. Hung. Acad. Sci}, vol.~5, no.~1, pp. 17--60, 1960.

\bibitem{watts1998collective}
D.~J. Watts and S.~H. Strogatz, ``Collective dynamics of
  ‘small-world’networks,'' \emph{nature}, vol. 393, no. 6684, pp. 440--442,
  1998.

\end{thebibliography}

\end{document}